\pdfoutput=1

\documentclass[11pt]{article}

\usepackage[]{acl}
\usepackage{nert}

\usepackage{times}
\usepackage{inconsolata}
\usepackage{multirow}
\usepackage{latexsym}
\usepackage{graphicx}

\usepackage[T1]{fontenc}

\usepackage[utf8]{inputenc}

\usepackage{microtype}
\usepackage{upgreek}
\usepackage{amsfonts}
\usepackage{amssymb}
\usepackage{amsmath}
\usepackage{cleveref}
\usepackage{mathtools}
\usepackage{algpseudocode}
\usepackage{float}
\usepackage{hyperref}
\newfloat{algorithm}{t}{lop}

\title{Syntactic Inductive Bias in Transformer Language Models:\\ Especially Helpful for Low-Resource Languages?}

\author{Luke Gessler \quad Nathan Schneider \\
  Department of Linguistics \\
  Georgetown University \\
  \{\emldisplay{lg876@georgetown.edu}{lg876}, \emldisplay{nathan.schneider@georgetown.edu}{nathan.schneider}\}\texttt{@georgetown.edu} \\}

\begin{document}
\maketitle

\begin{abstract}
A line of work on Transformer-based language models such as BERT 
has attempted to use syntactic inductive bias to enhance the pretraining process, 
on the theory that building syntactic structure into the training process should reduce the amount of data needed for training.
But such methods are often tested for high-resource languages such as English. 
In this work, we investigate whether these methods can compensate for data sparseness in low-resource languages, hypothesizing that they ought to be more effective for low-resource languages.
We experiment with five low-resource languages: Uyghur, Wolof, Maltese, Coptic, and Ancient Greek.
We find that these syntactic inductive bias methods produce uneven results in low-resource settings, and provide surprisingly little benefit in most cases.
\end{abstract}

\def\CodeRepo{
\url{https://github.com/lgessler/lr-sib}
}

\def\wordtovec{\textsc{word2vec}}
\def\mbert{\textsc{mbert}}
\def\mbertva{\textsc{mbert-va}}
\def\m{$\upmu$\textsc{b-m}}
\def\mp{$\upmu$\textsc{b-mp}}
\def\mt{$\upmu$\textsc{b-mt}}
\def\mpt{$\upmu$\textsc{b-mpt}}
\def\mptsla{$\upmu$\textsc{b-mpt-sla}}
\def\mx{$\upmu$\textsc{b-mx}}
\def\mxp{$\upmu$\textsc{b-mxp}}
\def\mxt{$\upmu$\textsc{b-mxt}}
\def\mxpt{$\upmu$\textsc{b-mxpt}}
\def\mxptsla{$\upmu$\textsc{b-mxpt-sla}}
\def\np{\textsc{-np}}
\def\pp{\textsc{-pp}}
\def\hqp{\textsc{-hqp}}
\def\bd{\textsc{-bd}}
\def\bm{\textsc{-bm}}

\section{Introduction}
Many NLP algorithms rely on high-quality pretrained word representations for good performance.
Pretrained Transformer language models (TLMs) such as BERT/mBERT \citep{devlin_bert_2019}, RoBERTa \citep{liu_roberta_2019}, XLM-R \citep{conneau_unsupervised_2020}, and ELECTRA \citep{clark_electra_2020} provide state-of-the-art word representations for many languages. 
However, these models require on the order of tens of millions of tokens of training data in order to achieve a minimum of quality \citep{micheli_importance_2020,warstadt_learning_2020}, a data requirement that most languages of the world cannot practically satisfy. 

There are at least two basic approaches to addressing this issue.
The first, which is at least as old as BERT, exploits multilingual transfer to reduce the data requirements for any individual language. 
The second aims to reduce TLMs' data requirements by modifying their architectures and algorithms.
For example, \citet{gessler-zeldes-2022-microbert} more effectively train low-resource monolingual TLMs with as few as 500K tokens by reducing model size and adding supervised pretraining tasks with part-of-speech tags and syntactic parses.

We take up the latter direction in this work, looking specifically at whether the addition of syntactic inductive bias (SIB) during the pretraining procedure may help improve TLM quality in low-resource, monolingual settings.
Specifically, we examine two methods which have been proposed for high-resource settings: the two syntactic contrastive loss functions of \citet{zhang-etal-2022-syntax}, and the modified self-attention algorithm of \citet{li-etal-2021-improving-bert}, wherein a modified self-attention mechanism, restricted so that tokens may only attend to tokens that are syntactically ``local'', complements the standard self-attention mechanism.

At a high level, SIB is of interest in the context of TLMs because of how crucial self-attention is for TLMs' syntactic knowledge.
In studies on an English TLM, BERT, \citet{htut2019} and \citet{clark-etal-2019-bert} show that while syntactic relations are not directly recoverable from self-attention patterns, many self-attention heads seem to be sensitive to particular syntactic relations, such as that of a direct object or or a subject.
But self-attention is \textit{completely} unbounded: during pretraining, the model has to learn from scratch how to decide which other tokens in an input sequence a token should attend to.
We therefore observe that if SIB could be effectively applied, then presumably self-attention weights would converge more quickly and learn more effectively, since their behavior has been observed to be so heavily syntactic in nature.

Moreover, we expect that this effect would be greater for low-resource languages, where the comparative lack of data is known to hamper models' ability to form robust linguistic representations.
We find additional motivation for our interest in SIB given the nearly universal view held by linguists that the human mind does not start with the equivalent of a totally unconstrained self-attention mechanism: for example, psycholinguists such as \citet{hawkins_cross-linguistic_2014} have extensively documented processing-related constraints on syntax, and Generative linguists such as \citet{ross_constraints_1967} have observed that many syntactic constructions which might have been possible are in fact not attested in English or any other language, and postulate that these constructions are at least in some cases ``impossible'' because of biologically-determined properties of the human mind.
Our goal is therefore to give our models something like the constraints the human mind has in order to help them learn more effectively with less data.


We use a standard BERT-like TLM architecture as our base model, though we heavily reduce model size, following the results of \citet{gessler-zeldes-2022-microbert} which showed that this is beneficial in low-resource monolingual settings.
We pretrain TLMs for five low-resource languages---Wolof, Coptic, Maltese, Uyghur, and Ancient Greek---varying which SIB methods are used.
We then use Universal Dependencies (UD) \citep{nivre_universal_2016} syntactic parsing and WikiAnn \citep{pan-etal-2017-cross} named entity recognition as representative downstream tasks that allow us to assess the quality of our models.
Additionally, we evaluate our models using PrOnto \citep{gessler2023pronto}, a suite of downstream task datasets for low-resource languages.
We find that these SIB methods are not very effective in low-resource languages, with small gains in some tasks and degradations or no effects in others. 
This is surprising given the intuition that SIB ought to help more in low-resource settings, and we speculate that other methods for SIB may be more effective in low-resource settings.

We summarize our contributions as follows:
\begin{enumerate}
\item We conduct what is, to the best of our knowledge, the first work examining whether SIB is helpful for pretraining low-resource Transformer LMs.
\item We reimplement SynCLM \citep{zhang-etal-2022-syntax}, SLA \citep{li-etal-2021-improving-bert}, and MicroBERT \citep{gessler-zeldes-2022-microbert} in plain PyTorch and make it openly accessible.\footnote{Our code is publicly available at \CodeRepo{}.}
\item We present evidence from seven downstream evaluation tasks wherein the two SIB methods we examine are basically ineffective in our experimental settings, yielding only scattered and small gains.
\end{enumerate}

\section{Previous Work}

Pretrained word representations have been essential ingredients for NLP models for at least a decade, beginning with static word embeddings such as word2vec \citep{mikolov_distributed_2013,mikolov_efficient_2013}, GloVe \citep{pennington_glove_2014}, and fastText \citep{bojanowski_enriching_2017}.
Contextualized word representations \citep{mccann_learned_2018,peters_deep_2018,devlin_bert_2019} from Transformer-based \citep{vaswani_attention_2017} models have since overtaken them.

Throughout this period, high-resource languages have received the majority of attention, and although interest in low-resource settings has increased in the past few years, there remains a large gap (in terms of linguistic resources, pretrained models, etc.)\ between low- and high-resource languages \citep{joshi-etal-2020-state}.

\subsection{Multilingual Models} 
The first modern multilingual TLM was mBERT, trained on 104 languages \citep{devlin_bert_2019}.
mBERT and other models that followed it, such as XLM-R \citep{conneau_unsupervised_2020}, demonstrated that multilingual pretrained TLMs are capable of good performance not on just languages represented in their training data, but also in some zero-shot settings (cf.~\citealt{pires_how_2019,rogers_primer_2020-3}, among others).
But this is not without a cost: it has been shown \citep{conneau_unsupervised_2020} that when a TLM is trained on multiple languages, the languages compete for parameter capacity in the TLM, which effectively places a limit on how many languages can be included in a multilingual model before performance significantly degrades for some or all of the model's languages.
Indeed, the languages which had proportionally less training data in XLM-R's training set tended to perform more poorly \citep{wu_are_2020}.

A possible solution to this difficulty is to {\it adapt} pretrained TLMs to a given target language, rather than trying to fit the target language into an ever-growing list of languages that the model is pretrained on.
One popular method for doing this involves expanding the TLM's vocabulary with additional subword tokens (e.g.~BPE tokens for RoBERTa-style models), which has been observed to improve tokenization and reduce out-of-vocabulary rates \citep{wang_extending_2020,artetxe-etal-2020-cross,chau_parsing_2020,ebrahimi_how_2021}, leading to downstream improvements in model performance.
But these and other approaches struggle when a language is very far from any other language that a multilingual TLM was pretrained on.

Multilingual models like XLM-R which are trained on over 100 languages could be described as massively multilingual models.
A more recent trend is to train multilingual models on just a few to a couple dozen languages, especially in low-resource settings.
For example, \citet{ogueji_small_2021} train an mBERT on data drawn from 11 African languages, totaling only 100M tokens (cf.~BERT's 3.3B), and find that their model outperforms massively multilingual models such as XLM-R, presumably because the African languages in question were quite unrelated to most of the languages \mbox{XLM-R} was trained on.

\subsection{Monolingual Models}
There has been comparatively little work exploring pretraining monolingual low-resource TLMs from scratch, and this lack of interest is likely explainable by the fact that monolingual TLMs require copious training data in order to be effective.
Several studies have examined the threshold under which monolingual models significantly degrade, and all find that using standard methods, more data than is available in ``low-resource'' settings (definitionally, if we take ``low-resource'' to mean `no more than 10M tokens') is required in order to effectively train a monolingual TLM.
\Citet{martin_camembert_2020} find at least 4GB of text is needed for near-SOTA performance in French, and \citet{micheli_importance_2020} show further for French that at least 100MB of text is needed for ``well-performing'' models on some tasks. 
\Citet{warstadt_learning_2020} train English RoBERTa models on datasets ranging from 1M to 1B tokens and find that while models acquire linguistic features readily on small datasets, they require more data to fully exploit these features in generalization on unseen data.

\citet{gessler-zeldes-2022-microbert} is the only work we are aware of which attempts to develop a method for training ``low-resource'' (<10M tokens in training data) monolingual TLMs.
They extend the typical MLM pretraining process with multitask learning on part-of-speech tagging and UD syntactic parsing, and also radically reduce model size to 1\% of BERT-base, yielding fair performance gains on two syntactic evaluation tasks.
They find that their monolingual approach generally outperforms multilingual methods for languages that are not represented in the training set of a multilingual TLM (mBERT, in their study).

\subsection{Syntactic Inductive Bias}\label{sec:sib}
Other work has investigated the syntactic capabilities of TLMs, and whether these capabilities could be enhanced with additional inductive bias.
In an influential study, \citet{hewitt_structural_2019} find that structures that resemble undirected syntactic dependency graphs are recoverable from TLM hidden representations using a simple ``structural probe'', consisting of a learned linear transformation and a minimum spanning tree algorithm for determining tokens' syntactic dependents based on L2 distance.
\citet{DBLP:conf/iclr/KimCEL20} find similar results with a non-parametric, distance-based approach using both hidden representations and attention distributions. 
Both of these works attempt to find syntactic representations within a TLM without ever exposing a TLM to a human-devised representation.
The quality of the recovered trees is usually poor relative to those obtainable from a syntactic parser, though their quality is consistently higher than random baselines.

Some works have attempted to provide models with direct access to human-devised representations---e.g., a syntactic parse provided in the Universal Dependencies formalism, which may have been produced by a human or by an automatic parser.
\citet{zhou_limit-bert_2020} extend BERT by adding dependency and constituency parsing as additional supervised tasks during pretraining.
\citet{bai_pre-train_2021} assume that inputs are paired with parses, and use the parses to generate masks which restrict an ensemble of self-attention modules to attend only to syntactic children, parents, or siblings.
\citet{xu-etal-2021-syntax} use dependency parses to bias self-attention so that self-attention between tokens is weighted proportionally to the tokens' distance in the parse.
In this paper, we examine the methods of \citet{li-etal-2021-improving-bert} and \citet{zhang-etal-2022-syntax}, which we describe below.

In sum, there are very many ways in which one could encourage a TLM to either learn a human representation of syntax, or to come up with (or reveal) its own.
To our knowledge, none of the works on SIB have been examined in a low-resource TLM pretraining setting.

\section{Approach}
This work investigates whether methods for SIB that have succeeded in high-resource monolingual TLM pretraining settings could also be useful in analogous low-resource settings.
As we have seen, monolingual TLMs tend to have very poor quality when less than $\approx$10M tokens of training data are available for pretraining, and moreover, it has been observed that at least one dimension of this poor quality is models' inability to make grammatical generalizations without a large ($\approx$1B tokens, \citealt{warstadt_learning_2020}) pretraining dataset.
Since it is (almost definitionally) difficult to get more data in low-resource settings, it is especially important to find other ways of improving model quality. 
It is therefore worthwhile to examine whether supplying some kind of SIB could help a low-resource TLM form better linguistic representations.

As discussed in \cref{sec:sib}, there are many ways to introduce SIB into a TLM.
In this work, we look specifically at two methods: SynCLM \citep{zhang-etal-2022-syntax} and SLA \citep{li-etal-2021-improving-bert}, which is also used by Zhang et al.
\citet{li-etal-2021-improving-bert} extend the self-attention module with ``local attention'', wherein tokens may only attend to tokens which are $\leq k$ edges away in the dependency parse tree.
\citet{zhang-etal-2022-syntax} devise two contrastive loss functions which are intended to encourage tokens to attend to sibling and child tokens, and in their experiments, they find success in combining these with SLA.
A concise description of the details of each method is available in Appendix \ref{sec:methods}.
Both of these methods have only been evaluated on English, and both assume a UD syntactic parse as an additional input for each input sequence and use the parse in different ways to attempt to guide the model to better syntactic representations.

We use these two SIB methods with the model of \citet{gessler-zeldes-2022-microbert}, MicroBERT, as a foundation.
MicroBERT is a BERT-like model that has been scaled down to 1\% of BERT-base, and that optionally employs part-of-speech tagging and syntactic parsing as auxiliary pretraining tasks.
As shown by experiments on 7 low-resource languages conducted by \citet{gessler-zeldes-2022-microbert}, MicroBERT performs much better than an unmodified \mbox{BERT-base} TLM, so we adopt it as our baseline model for most experiments in this work.

We now state our two main research questions:
\begin{itemize}
    \item \textbf{(RQ1)} Do these SIB methods improve model quality when applied to a low-resource language?
    \item \textbf{(RQ2)} Are there any gains \textit{complementary} with the part-of-speech tagging component of MicroBERT for training low-resource monolingual TLMs?
\end{itemize}

\section{Methods}
\subsection{Data and Evaluation}
\begin{table}[]
    \centering
    \small
    \begin{tabular}{l|ccc}
Language   & Unlabeled & UD & NER\\\hline
Wolof      &  \hphantom{0,}517,237 &   \hphantom{00}9,581 &   10,800 \\
Coptic     &  \hphantom{0,}970,642 &   \hphantom{0}48,632 &       -- \\
Maltese    &  2,113,223 &   \hphantom{0}44,162 &   15,850 \\
Uyghur     &  2,401,445 &   \hphantom{0}44,258 &   17,095 \\
Anc.~Greek &  9,058,227 &  213,999 &       -- \\
    \end{tabular}
    \caption{Token count for each dataset by language from \citet{gessler-zeldes-2022-microbert}, sorted in order of increasing unlabeled token count.}
    \label{tab:token_stats}
\end{table}

We reuse the datasets and evaluation setup of \citet{gessler-zeldes-2022-microbert}, using five of their seven ``truly''\footnote{The Indonesian and Tamil Wikipedias were larger than Gessler and Zeldes' cutoff of 10M tokens for ``low resource'', and Indonesian and Tamil are also included in mBERT's pretraining data. We exclude them for the purposes of this study in the interest of examining these five truly low-resource languages in more depth.} low-resource languages' datasets. 
Each language's data includes a large collection of unlabeled pretraining data sourced from Wikipedia, as well as two datasets for downstream tasks for evaluation: UD treebanks for syntactic parsing, and WikiAnn \citep{pan-etal-2017-cross} for named entity recognition (NER).
We refer readers to Gessler and Zeldes' paper for further details on these datasets and the models for UD parsing and NER.
In addition, we assess models on all five tasks in the PrOnto benchmark \citep{gessler2023pronto}, which will be described below.

\begin{table*}[t]
    \centering
    \footnotesize
    \begin{tabular}{l|ccccc|c}
        Model                        & Wolof & Coptic & Maltese & Uyghur & An.~Gk. & Avg. \\\hline\hline
        \mbert                       & 76.40 & 14.43 & 78.18 & 46.30 & 72.30 & 57.52 \\
        \mbertva                     & 72.94 & 82.11 & 72.69 & 42.97 & 65.89 & 67.32 \\\hline
        \textsc{$\upmu$b-m}          & 77.71 & 88.47 & 81.40 & 59.97 & 81.94 & 77.90 \\
        \textsc{$\upmu$b-mp}         & 75.88 & 87.90 & 80.88 & 59.42 & 81.15 & 77.05 \\
        \textsc{$\upmu$b-mt}         & 77.29 & 88.32 & 81.06 & 59.79 & 81.42 & 77.58 \\
        \textsc{$\upmu$b-mpt}        & 77.05 & 88.38 & 80.07 & 58.94 & 81.35 & 77.16 \\
        \textsc{$\upmu$b-mpt-sla}    & 76.25 & 87.87 & 79.52 & 58.37 & 80.77 & 76.56 \\\hline
        \textsc{$\upmu$b-mx}         & 77.74 & 88.00 & 81.25 & 61.23 & 82.02 & 78.05 \\
        \textsc{$\upmu$b-mxp}        & 77.90 & 88.63 & 82.21 & 60.62 & 81.34 & 78.14 \\
        \textsc{$\upmu$b-mxt}        & 77.30 & 88.34 & 81.87 & 60.44 & 82.11 & 78.01 \\
        \textsc{$\upmu$b-mxpt}       & 78.19 & 88.48 & 81.30 & 61.41 & 81.80 & 78.24 \\
        \textsc{$\upmu$b-mxpt-sla}   & 76.89 & 87.90 & 80.87 & 59.35 & 81.17 & 77.24 \\
    \end{tabular}
    \caption[Main UD parsing results for SynCLM/SLA]{Labeled attachment score (LAS) by language and model combination for UD parsing evaluation. Results for \mbert{} and \mbertva{} are taken from \citet{gessler-zeldes-2022-microbert}.}
    \label{tab:loreiba_parser_results}
\end{table*}

\begin{table}[t]
    \centering
    \footnotesize
    \begin{tabular}{l|ccc|c}
        Model                        & Wolof & Maltese & Uyghur & Avg. \\\hline\hline
        \mbert                       & 83.79 & 73.71 & 78.40 & 78.63 \\
        \mbertva                     & 79.37 & 78.11 & 77.03 & 78.17 \\\hline
        \textsc{$\upmu$b-m}          & 83.40 & 82.98 & 86.70 & 84.36 \\
        \textsc{$\upmu$b-mp}         & 86.38 & 84.16 & 87.44 & 86.00 \\
        \textsc{$\upmu$b-mt}         & 87.16 & 89.46 & 87.33 & 87.98 \\
        \textsc{$\upmu$b-mpt}        & 88.89 & 86.83 & 87.67 & 87.80 \\
        \textsc{$\upmu$b-mpt-sla}    & 86.38 & 84.85 & 84.81 & 85.35 \\\hline
        \textsc{$\upmu$b-mx}         & 77.65 & 86.09 & 89.75 & 84.49 \\
        \textsc{$\upmu$b-mxp}        & 81.45 & 87.74 & 87.41 & 85.54 \\
        \textsc{$\upmu$b-mxt}        & 85.94 & 84.67 & 87.98 & 86.19 \\
        \textsc{$\upmu$b-mxpt}       & 87.06 & 84.37 & 87.53 & 86.32 \\
        \textsc{$\upmu$b-mxpt-sla}   & 83.72 & 85.35 & 88.07 & 85.71 \\
    \end{tabular}
    \caption[Main NER results for SynCLM/SLA]{Span-based F1 score by language and model combination for NER evaluation.}
    \label{tab:loreiba_ner_results}
\end{table}

\subsection{Models}
We reimplement the MicroBERT model of \citet{gessler-zeldes-2022-microbert}, as well as the work of \citet{zhang-etal-2022-syntax} and \citet{li-etal-2021-improving-bert}.
In all cases, we reuse code wherever possible and closely check implementation details and behavior in order to ensure correctness.
As a foundation, we use the BERT implementation provided in HuggingFace's \texttt{transformers} package \citep{wolf-etal-2020-transformers}, and we also use AI2 Tango\footnote{\url{https://github.com/allenai/tango}} for running experiments.
We obtain all of our parses for the unlabeled portions of our datasets automatically using Stanza \citep{qi-etal-2020-stanza}, following Zhang et al.

In order to answer our research questions, for each language, we examine the following conditions:
\begin{enumerate}
\setlength\itemsep{0em}
    \item \mbert{} -- plain multilingual BERT (\texttt{\small bert-base-multilingual-cased}). A baseline; numbers taken from Gessler and Zeldes.
    \item \mbertva{} -- \mbert{}, but with vocabulary augmentation. A baseline; numbers taken from Gessler and Zeldes.
    \item \m{} -- plain MicroBERT trained only using MLM. We obtain our own numbers to verify the correctness of our implementation.
    \item \mp, \mt, \mpt{} -- MicroBERT with either one or both of the SynCLM loss functions: \textsc{p} indicates the phrase-guided loss, and \textsc{t} indicates the tree-guided loss.
    \item \mptsla{} -- \mpt{}, with the addition of SLA. We follow \citet{zhang-2022-improve} in using SLA only in conjunction with both contrastive losses.
    \item \mx, \mxp, \mxt, \mxpt, \mxptsla -- the conditions in (3--5), but with the addition of part-of-speech tagging (\textsc{x}) as an auxiliary pretraining task. This is done using the same methods of Gessler and Zeldes: PoS tagging is only performed on gold-tagged data from the UD treebank, and tagged sequences are mixed into the pretraining data at a 1 to 8 ratio.
\end{enumerate}

Revisiting our research questions, we intend for the conditions in (3--5) to provide evidence for \textbf{(RQ1)}, and for the additional information from the conditions in (6) to provide evidence for \textbf{(RQ2)}.

\section{Results}

\begin{table*}[t]
    \centering
    \small
    \begin{tabular}{@{}l|cccc|c||cccc|c||c}
                                     & \multicolumn{5}{c}{\textbf{Non-pronominal Mention Count}} & \multicolumn{5}{c}{\textbf{Same Sense}} & \textbf{All 5}\\
        Model                        & An. Grk. & Coptic   & Uyghur & Wolof & Avg. & An. Grk. & Coptic & Uyghur & Wolof & Avg.   & Avg. \\\hline
        \textsc{$\upmu$b-m*}         & 52.59    & 50.75    & 49.37  & 51.47 & 51.04 & 60.58    & 61.32  & 60.65  & 59.78 & 60.58 &   68.65\\
        \textsc{$\upmu$b-mx*}        & 56.81    & 53.34    & 51.19  & 59.24 & 55.14 & 60.95    & 61.30  & 61.51  & 63.08 & 61.71 & 70.01  \\
        \mbert                       & 57.36    & 49.52    & 51.46  & 57.35 & 53.92 & 65.34    & 52.79  & 62.73  & 66.49 & 61.84 
 & 67.92 \\\hline
        \textsc{$\upmu$b-m}          & \underline{56.68}    & \underline{52.52}    & 52.72  & 53.78 & 53.93 & \textbf{\underline{58.51}}    & 56.65  & 57.97  & 58.54 & 57.92 & 68.25 \\
        \textsc{$\upmu$b-mp}         & 56.13    & 51.98    & \underline{\textbf{54.39}}  & \underline{\textbf{54.41}} & \underline{54.23} & 58.41    & \underline{\textbf{58.15}}  & \underline{59.54}  & \underline{\textbf{58.95}} & \textbf{\underline{58.76}} & \textbf{\underline{68.40}}  \\
        \textsc{$\upmu$b-mt}         & 50.41    & 48.98    & 49.37  & 51.47 & 50.06 & 58.48    & 58.08  & 57.99  & 57.03 & 57.90 & 66.88  \\
        \textsc{$\upmu$b-mpt}        & 53.68    & 48.98    & 51.74  & 51.47 & 51.47 & 53.36    & 54.19  & 59.32  & 58.07 & 56.23 & 66.39 \\\hline
        \textsc{$\upmu$b-mx}         & \textbf{\underline{57.49}}    & 53.07    & \underline{\textbf{54.39}}  & 53.57 & \textbf{\underline{54.63}} & 56.71    & 56.01  & 58.88  & 58.18 & 57.44 & \underline{68.39}  \\
        \textsc{$\upmu$b-mxp}        & 54.09    & \underline{\textbf{53.34}}    & \underline{\textbf{54.39}}  & \underline{53.78} & 53.90 & 55.61    & 55.02  & 59.47  & \underline{58.47} & 57.14 & 67.84   \\
        \textsc{$\upmu$b-mxt}        & 53.95    & 51.02    & 49.37  & 51.47 & 51.45 & \underline{57.44}    & \underline{56.37}  & \underline{\textbf{59.56}}  & 57.93  & \underline{57.83} & 66.89 \\
        \textsc{$\upmu$b-mxpt}       & 52.72    & 51.71    & 50.91  & 51.47 & 51.70 & 57.19    & 56.17  & 56.81  & 58.14 & 57.08 & 67.30  \\
    \end{tabular}
    \caption{Accuracy by language and model combination for two tasks in PrOnto: the Non-pronominal Mention Count, and Same Sense tasks. For non-baseline models, an underline indicates the best performance for a language--task combination for a particular model variant (\textsc{-m} or \textsc{-mx}), and boldface indicates the best performance across either model variant. Scores for \mbert{}, \textsc{$\upmu$b-m*}, and \textsc{$\upmu$b-mx*} are taken from \citet{gessler2023pronto}---the asterisk indicates that the latter two models are not our implementation but the one provided in \citet{gessler-zeldes-2022-microbert}, which is reported in \citet{gessler2023pronto}. Rightmost column contains an average over all languages and tasks for a given model. Results for PrOnto's other three tasks are given in Appendix \ref{sec:loreiba_pronto_extra_results}.}
    \label{tab:loreiba_pronto_results}
\end{table*}

\paragraph{Parsing} 
Our results for UD syntactic parsing are given in Table \ref{tab:loreiba_parser_results}.
While all models beat the multilingual baselines, neither SynCLM nor SLA seems to improve model quality.
In the \textsc{-m} variant models, the top-performing model is always the one trained with plain masked language modeling.
This is not so for the \textsc{-mx} variant models, where the \textsc{-mxp} and \textsc{-mxpt} models do slightly better on average, though this difference is small enough to be within the range of experimental noise.
Surprisingly, \textsc{-mpt-sla} models do worst of all.
Finally, comparing \textsc{-m} variants to their \textsc{-mx} counterparts, we do find that in all cases the \textsc{-mx} counterpart is better on average, and that the difference is about 1\% LAS.

\paragraph{NER}
Our results for WikiAnn NER are given in Table \ref{tab:loreiba_ner_results}.
Considering the \textsc{-m} variant models first, we see that in all cases the model trained using only MLM performs the worst, and the \textsc{-mpt-sla} variant, while always no better than the \textsc{-mp}, \textsc{-mt}, and \textsc{-mpt} variants, also outperforms the plain MLM model.
The \textsc{-mp}, \textsc{-mt}, and \textsc{-mpt} variants do best with a difference of up to 4 points F1 on average.

Turning now to the \textsc{-mx} variants, while it is still true that on average the plain MLM model performs worst and the non-SLA SynCLM models perform best, there is more variation within individual languages.
The best model for Uyghur is the plain MLM model, and for Maltese, the plain MLM model outperforms \mxt{} and \mxpt.

Considering now all the NER results, two patterns are worth noticing.
First, unlike in parsing, a \textsc{-mx} variant does not always outperform its \textsc{-m} counterpart: for example, \mp{} for Wolof is better than \mxp{} by a difference of 5 points F1.
We can see further that the \textsc{-m} models beat the \textsc{-mx} models on average by about 4 points F1. 
This indicates that when combined with SLA and SynCLM, the PoS tagging pretraining task does not appear to be helpful for dimensions of model quality that are implicated in NER.
Second, the addition of \mbox{\textsc{-sla}} never results in a gain relative to any of the SynCLM models, except for Uyghur, where it produces a gain of 0.09, which is within the range of experimental noise.

\paragraph{PrOnto}

We run our SynCLM models\footnote{It was not possible to run our SLA models on PrOnto due to considerable implementation effort that would have been required, so we omit those models from this evaluation.} on all five tasks of PrOnto \citep{gessler2023pronto} on all languages except Maltese, which is not represented in PrOnto because of the lack of an open-access Maltese Bible.
For each language in PrOnto, a dataset for five sequence classification tasks is available which was constructed by aligning New Testament verses from the target language with the English verse in OntoNotes \citep{hovy_ontonotes_2006} and projecting annotations from English to the target language. 
All 5 tasks are sequence classification tasks.
Each task requires a model to predict a certain grammatical or semantic property---these are, respectively: the number of referential noun phrases in a sequence; whether the subject of a sentence contains a proper noun; the sentential mood of a sentence; whether two input sequences both contain a usage of a verb sense; and whether two input sequences both contain a usage of a verb sense with the same number of arguments.
We refer readers to the PrOnto publication for further details.

Results from two of the five tasks are given in Table \ref{tab:loreiba_pronto_results}.\footnote{We omit results from the other 3 from the main body for space reasons---see Appendix \ref{sec:loreiba_pronto_extra_results} for these results.}
Broadly, we may observe that the \textsc{-mpt} and \textsc{-mxpt} models never perform best within a language, with either variant being in many cases worse by a few absolute points compared to other models.
Looking at \textsc{-m}-family models, \textsc{-mp} is the clear winner, doing a little better than \textsc{-m} and much better than \textsc{-mt} or \textsc{-mpt} on both tasks.
By contrast, for \textsc{-mx}-family models, the \textsc{-mxp} variant does a bit worse on average than \textsc{-mx}, and for the Same Sense task, the \textsc{-mxt} model does a bit better than \textsc{-mxp}. 
Looking to the rightmost column in Table \ref{tab:loreiba_pronto_results}, we can see that when we average accuracy scores for a model across all languages and all 5 tasks in PrOnto, the \textsc{-mp} model has the highest score overall, with \textsc{-mx} and \textsc{-m} very close behind and all other model variants quite a ways behind.

Overall, it seems that for the PrOnto tasks, of all the syntactic bias methods we have tried, only the use of the phrase-based contrastive loss (\textsc{-mp}) or the tree-based contrastive loss in combination with PoS tagging (\textsc{-mxt}) showed much improvement over the baselines.
In individual language--task combinations, models sometimes had multiple-point performance differences over others, but when considered in aggregate, only \textsc{-mp} shows any improvement over \textsc{-m} and \textsc{-mx}---by 0.15\% and 0.01\% accuracy, respectively.

\section{Discussion}
Considering first whether SynCLM and SLA yield benefits for low-resource monolingual TLMs (RQ1), we have found positive evidence from the WikiAnn NER experiments, and weak positive evidence from the PrOnto experiments.
It is true that the same methods did not produce measurable gain for the UD parsing task, but this is in line with previous findings for these two methods, where on some downstream evaluations, gain was very small or slightly negative---we return to this matter in the following paragraph.
For the question of whether these benefits are complementary with the PoS tagging pretraining strategy introduced in \citet{gessler-zeldes-2022-microbert} (RQ2), we do not find consistent evidence in any of our experiments that both PoS tagging and SynCLM or SLA yield complementary benefits.
The only positive evidence we find for this is in the PrOnto experiments, where the \textsc{-mxt} model variant does better than \textsc{-mx} in some task--language combinations, though worse overall.

The difference in the way model variants behaved in these seven evaluation tasks is striking, and it is difficult to understand why models exhibited these different behaviors.
It is worth comparing these results with those reported by the SynCLM authors \citep{zhang-etal-2022-syntax}.
For many of the GLUE tasks that they assess their models on (their Table 3), there is little or no improvement from adding \textsc{-p}, \textsc{-t}, or \textsc{-pt-sla}.
For example, considering their models based on RoBERTa-base, none of their model variants outperform the MLM-only baseline for the QQP (Quora Question Pairs2), STS (Semantic Textual Similarity), or MNLI-m (Multi-Genre Natural Language Inference, matched).
This situation is more or less analogous to the one we observed in our experiments for the UD parsing downstream task, where the addition of SynCLM and SLA had basically no effect.

On the other hand, the GLUE task with the greatest gain, CoLA (Corpus of Linguistic Acceptability), shows a difference of only 1.7\% Matthews correlation coefficient, and a couple of other tasks like SST (Stanford Sentiment Treebank), show an improvement of only 0.3\% accuracy.
It would be na\"ive to directly compare percentage points of different metrics in totally different experimental settings and make conclusions about effect sizes, we nevertheless point out that we observe improvements of 1--4\% F1 in our NER experiments for \textsc{-m} models.
In light of this, we consider our results to be broadly in line with the trend for previous works' results on English: there is no improvement that is wholly consistent across evaluations, and only modest gains for the benchmarks that do improve.

In summary, we find that SynCLM and SLA produce uneven results in low-resource settings, though we also find that when they do succeed, they can yield gains that appear greater than anything observed for high-resource languages: we saw that when we take a pure MLM pretraining regimen as a base and add SynCLM and/or SLA, we are able to improve the quality of pretrained TLMs by 1 to 4 absolute points F1 in NER.
While a similar benefit was not observed for UD parsing, it is also true that there was a noticeable degradation on UD parsing in only a couple cases, and in most cases simply had no effect.


\section{English Experiments}
One might have expected SIB to be a knockout success for low-resource languages given the intuitive feeling that at lower data volumes, additional bias ought to be more helpful.
We considered reasons why our attempts to do this might not have panned out---perhaps, for example, tree structure matters most for highly analytic languages like English, or perhaps the tasks used to evaluate English in GLUE are more sensitive to high-level sentence structure, or perhaps sensitivity to syntax is only advantageous given a base model with sufficiently rich distributional information.
Here, we consider another possible explanation: that the inductive bias with these methods only helps given high-quality syntactic parses.
An obvious difference between English and the languages we have examined in this study is that UD parsers for English generally achieve much higher performance given the size and annotation quality of English UD treebanks.
This is a potentially consequential difference, given that both the SynCLM and SLA methods rely on UD parse trees as inputs.
In addition, the models we have developed here differ from common kinds of English BERTs in that they are much smaller and were trained on much less data, and it is possible that the SynCLM and SLA methods might have interactions with these two variables of model construction.

In order to investigate whether parse tree quality, model size, and pretraining data size might be consequential for these SIB methods, we run several additional experiments on English datasets.
We choose English because its status as a high-resource language allows us control over several independent variables which we do not have control over in low-resource settings, namely data quantity, syntactic parse quality, and model size\footnote{Model size is not controllable in low-resource settings in the sense that, as \citet{gessler-zeldes-2022-microbert} argued, monolingual low-resource TLMs exhibit severe degradations when they get too large.}.
We can frame an additional research question that we wish to answer:
\begin{itemize}
    \item \textbf{(RQ3)} Are SynCLM and SLA sensitive to parse tree quality, model size, or pretraining dataset size?
\end{itemize}

For our English dataset, we use AMALGUM \citep{gessler-etal-2020-amalgum} as our source of pretraining data.
AMALGUM contains around 2M tokens and contains automatic parses with quality that exceeds what can normally be obtained from a standard parser.
For downstream evaluation, we use the English Web Treebank \citep{silveira14gold}, which contains around 250K tokens, and the English split of WikiAnn, downsampled to around 50K tokens in order to bring it closer to the quantities for our other 3 languages (cf.~Table \ref{tab:token_stats}).
In addition, we use a 100M subset of BERT's pretraining data as a larger source of unlabeled pretraining data.

We frame these additional conditions for English, extending our model naming scheme from above:
\begin{enumerate}
\setlength\itemsep{0em}
    \item \np{} -- syntax trees are taken from Stanza in the same way as before.
    \item \hqp{} -- syntax trees are taken from AMALGUM's annotations, made by a \textbf{h}igh \textbf{q}uality \textbf{p}arser.
    \item \bd{} -- pretraining is done using the \textbf{b}ig \textbf{d}ataset instead of AMALGUM.
    \item \bd{}\bm{} -- like \bd{}, and in addition, the model size is set to half of BERT-base (6 layers instead of 12).
\end{enumerate}
Evidence from these conditions could tell us more about how and when SynCLM and SLA can succeed in low-resource scenarios.
We pretrain these models as we did in our main experiments and evaluate them on UD parsing and WikiAnn NER.

A full description of our results is given in Appendix \ref{sec:english}, and we give a description of our key finding here: that SynCLM and SLA are not very sensitive to parse quality or model size, but are sensitive to quantity of pretraining data.
The insensitivity to parse quality may come as a surprise, and we reason that this is actually understandable, since both methods focus mostly on low-height subtrees (often corresponding to phrase- or sub-phrase-level constituents) which are more likely to be correct even when overall parse quality is bad.
We find evidence for sensitivity to data size in the fact that SynCLM and SLA provide gains of up to 1\% F1 for the NER evaluation in the two low-data conditions, while in the higher-data conditions, all but one of the bias-enhanced models lead to degradations relative to the baseline.
In sum, we take this to show that lower parse quality is not the major reason for the ineffectiveness of SynCLM and SLA in low-resource settings.

\section{Conclusion}
In this work, we have taken two methods for SIB that have succeeded in English, SynCLM and SLA, and we have investigated whether they may also be beneficial in low-resource monolingual settings.
We find that in most cases these methods do not result in an improvement in model quality as measured on seven tasks.
Further, in our auxiliary experiments on English, we found evidence suggesting that the lower quality of parses in low-resource settings is probably not what is driving the ineffectiveness of these SIB methods.

Considering all of our results, we conclude that these two specific methods---SynCLM and SLA---are not well suited to supporting the pretraining of language models in low-resource settings, but we also view it as a yet open question whether any method for SIB could succeed in this role.
There are some reasons why SynCLM and SLA might have been unhelpful.
First of all, recall the fact that SynCLM limits its application to only short subtrees (no taller than 3 nodes).
This would mean that most of the time, the contrastive loss functions would only be operating on basic phrase-level constituents, such as noun phrases, and not higher, clause-level phenomena such as relations between the main clause's predicate and its arguments.
If it were the case that the former kind of syntax is relatively easy for models to learn even with limited data, and that the latter kind of syntax is what is hard and therefore where SIB really ought to help, then we would expect to see the results we found in this work, where neither method did much to help.

Therefore, while we find little reason to be optimistic about these two particular methods in low-resource settings, we don't view the evidence in this paper as an indictment of SIB in low-resource settings in general, and suggest that SIB methods which are better able to provide bias for higher, clause-level syntactic dependencies may produce better results for low-resource languages.



\section*{Acknowledgments}
We thank Amir Zeldes for very helpful comments on this work.

\bibliography{anthology,custom}
\bibliographystyle{acl_natbib}

\clearpage
\appendix

\section{Summary of SLA and SynCLM}
\label{sec:methods}
Our approach critically relies on two previous results, which we summarize here.

\subsection{Syntax-aware Local Attention}\label{sec:sla}
\begin{figure}
    \centering
    \includegraphics[width=\linewidth]{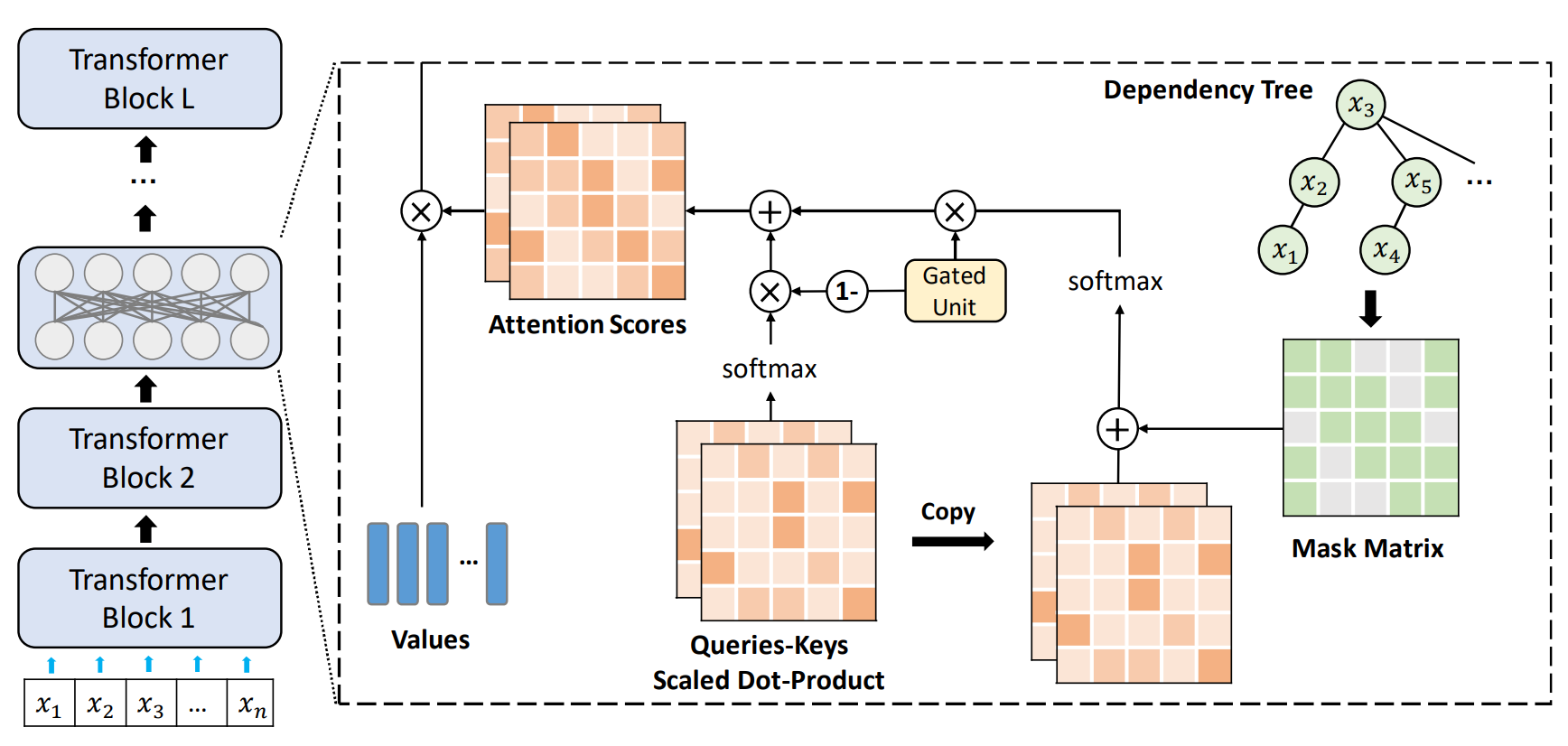}
    \caption{Figure 1 from \citet{li-etal-2021-improving-bert}. The standard self-attention mechanism is complemented by another self-attention mechanism in which tokens may only attend to tokens close to it in a parse tree. A gated unit with learnable parameters interpolates the two attention distributions before the distribution is combined with the Value representation.}
    \label{fig:sla}
\end{figure}

\citet{li-etal-2021-improving-bert} introduce Syntax-aware Local Attention (SLA), a variation on a standard TLM self-attention mechanism that retains standard self-attention and complements it with a separate self-attention mechanism where each token may only attend to ``syntactically local'' tokens.

Recall that BERT and most other TLMs use scaled dot-product attention in every attention head, where the attention distribution $\mathbf{A}$ can be computed with query and key representations $\mathbf{Q}$ and $\mathbf{K}$, $d$ is the size of an individual attention head's hidden representation, and the attention head's output $\mathbf{O}$ is the product of $\mathbf{A}$ and the value representation $\mathbf{V}$:
\begin{equation}\label{eq:attention}
    \mathbf{A} = \mathrm{softmax}\left(\frac{\mathbf{QK}^\top}{\sqrt{d}}\right)
\end{equation}
\begin{equation}
    \mathbf{O} = \mathbf{A}\mathbf{V}
\end{equation}

Now, assume an input sequence $W = w_1, \ldots, w_n$ with an unlabeled dependency parse $H = h_1, \ldots, h_n$ where $h_i$ indexes token $w_i$'s syntactic head.
Define syntactic distance between two words, $D(w_i, w_j)$, as the length of the shortest path between the two words in the parse:
\begin{equation}
    D(w_i, w_j) \coloneqq \textsc{Shortest-Path}(H, i, j)
\end{equation}
To account for the fact that parses may be inaccurate (e.g.~if they come from an automatic parser), define \textit{windowed} syntactic distance like so:\footnote{If $k \not\in [1, n]$, exclude it from the $\min$.}
\begin{equation}\label{eq:windowed}
    D^\prime(w_i, w_j) = \min_{k \in \{i - 1, i, i + 1\}} D(w_k, w_j)
\end{equation}
This can be viewed as sacrificing precision for recall: a decision to give tokens a better chance of being able to attend to truly local tokens (given the imperfection of parser outputs), though at the cost of sometimes allowing attention on tokens that truly are not local.

Now, define a mask matrix $\mathbf{M}$ that will mask a token \textit{iff} a token $j$ has windowed syntactic distance over a certain threshold $\delta$ relative to token $i$:
\begin{equation}
    m_{ij} = \begin{cases}
    0 & \mathrm{if}\ D^\prime(w_i, w_j) \leq \delta \\
    -\infty & \mathrm{otherwise} 
    \end{cases}
\end{equation}
We can now define syntax-aware local attention by modifying Equation \ref{eq:attention} so that $\mathbf{M}$ is added to the inner term in order to force an attention score of 0 for masked tokens:
\begin{equation}
    \mathbf{A^\ell} = \mathrm{softmax}\left(\frac{\mathbf{QK}^\top}{\sqrt{d}} + \mathbf{M} \right)
\end{equation}

Syntax-aware local attention (SLA) is used alongside the normal, ``global'' self-attention.
To combine the two after they have been computed, introduce a gated unit for each Transformer block with new parameters $\mathbf{W}_g$ and $b_g$ to compute $g_i$ for each word $w_i$ using the word's hidden representation $\mathbf{h}_i$, where $\sigma$ is the sigmoid function:
\begin{equation}
    g_i = \sigma(\mathbf{W}_g\mathbf{h}_i + b_g)
\end{equation}
Now, use $g_i$ to interpolate both the normal attention distribution $\mathbf{a}_i$ and the local attention distribution $\mathbf{a}^\ell_i$ at each position $i$ in the sequence to yield the final attention distribution $\mathbf{\hat{A}}$ and final attention head output $\mathbf{\hat{O}}$:
\begin{equation}
    \mathbf{\hat{A}} = \bigoplus_{i=1}^n g_i\mathbf{a}_i + (1 - g_i)\mathbf{a}^\ell_i
\end{equation}
\begin{equation}
    \mathbf{\hat{O}} = \mathbf{\hat{A}}\mathbf{V}
\end{equation}

In the original work, the SLA method is evaluated on various benchmarks on English and consistently achieves measurable improvements in model quality.
Parses are obtained using Stanza \citep{qi-etal-2020-stanza}, which 
for English are of quite high quality (labeled attachment score is in the mid-80s for English datasets).
We refer readers to the original publication for further details.
See Figure \ref{fig:sla} for an overview.

\subsection{SynCLM}
\begin{figure*}
    \centering
    \includegraphics[width=\linewidth]{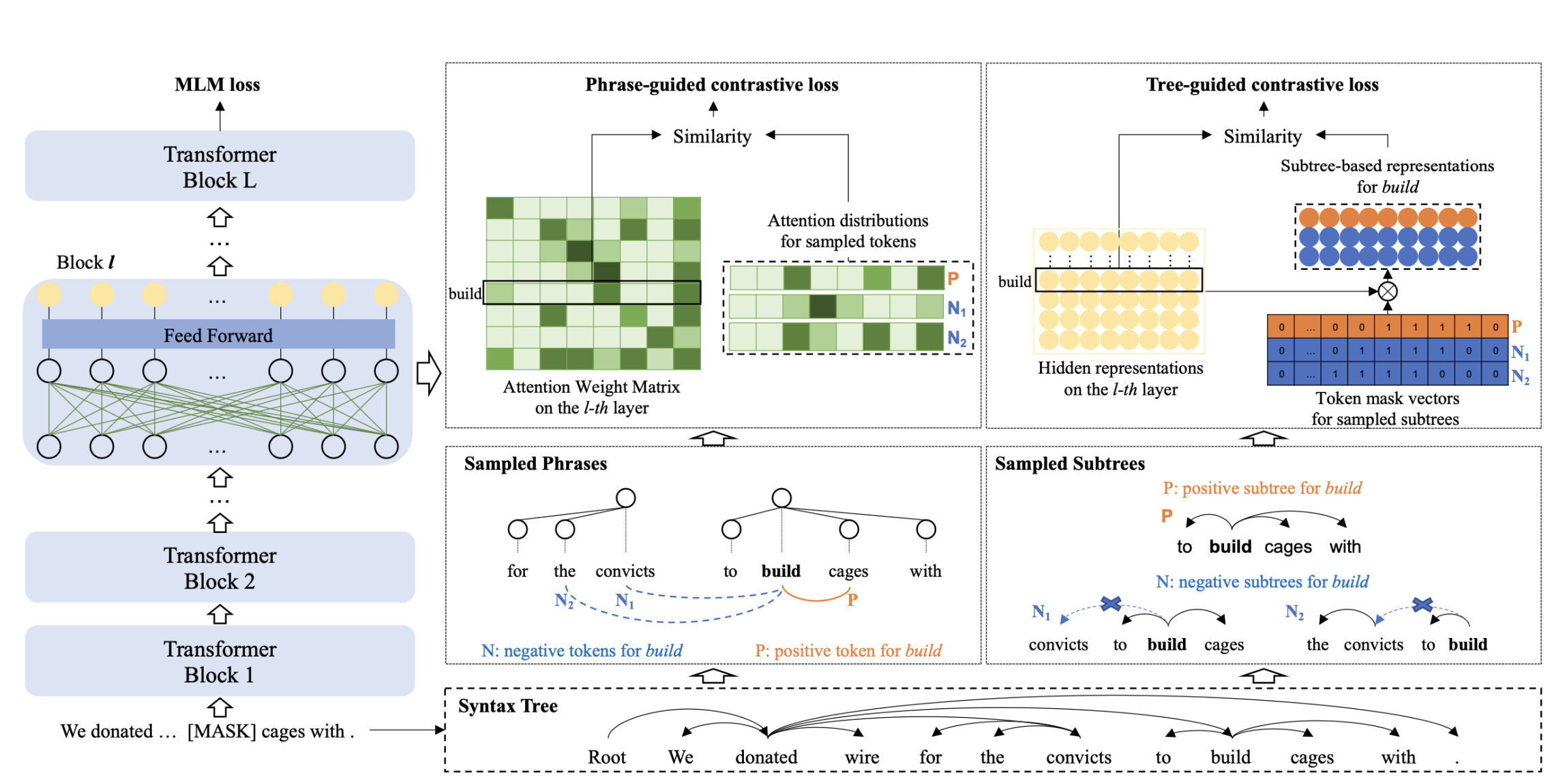}
    \caption{Figure 1 from \citet{zhang-etal-2022-syntax}. \textbf{\textit{P}} and \textbf{\textit{N}}$_i$ represent the positive sample and the $i$th negative sample, respectively. The phrase-based contrastive loss on the left is intended to make the representations of syntactic siblings more similar, and the tree-based contrastive loss on the right is intended to make the representations of syntactic children and parents more similar.}
    \label{fig:synclm}
\end{figure*}

\citet{zhang-etal-2022-syntax} present the Syntax-guided Contrastive Language Model (SynCLM), a BERT-like TLM that characteristically uses two novel contrastive loss functions and also uses SLA (cf.~\cref{sec:sla}).
Intuitively, a contrastive learning objective requires each instance to have one or more \textit{positive} and \textit{negative} ``samples'', and attempts to maximize the instance's similarity to positive samples and minimize its similarity to negative samples \citep{zhang-etal-2022-contrastive-data}.
SynCLM uses a popular loss function for this, InfoNCE \citep{infonce}:
\begin{equation}\label{eq:infonce}
\small
    \mathrm{L} = -\log 
    \frac
    {\exp\left(\frac{\mathrm{sim}(q,q^+)}{\tau}\right)}
    {\exp\left(\frac{\mathrm{sim}(q,q^+)}{\tau}\right) + \sum_{i=0}^K \exp\left(\frac{\mathrm{sim}(q,q_i^-)}{\tau}\right)}
\end{equation}
$q$, $q^+$, and $q^-$ are the representations of the instance, a positive sample, and a negative sample, respectively, and $\tau \in (0, 1)$ is a temperature hyperparameter, set to $0.1$ for SynCLM.
$\mathrm{sim}$ is a similarity function, such as cosine similarity or KL-divergence.
The loss terms obtained from this equation are simply added to the loss obtained from masked language modeling.
We review only the contrastive objective functions here, and refer readers to Figure \ref{fig:synclm} and the original paper for further details.

The two SynCLM contrastive learning objectives are distinguished by how they formulate $\mathrm{sim}$.
The first, ``phrase-guided'' objective aims to make attention distributions more similar for words in the same phrase.
Given a token $t$, sample a positive token $t^+$ such that $t$ and $t^+$ have a lowest common ancestor $t_a$ whose corresponding subtree (the ``phrase'') is no more than 2 in height.
Now sample $k$ negative tokens $t^-_1, \ldots, t^-_k$ outside the phrase, i.e.~who do not have $t_a$ as an ancestor. 
Define $\mathrm{sim}_\mathrm{phrase}$ using Jensen--Shannon Divergence \citep{endres2003new}, a similarity metric for probability distributions:
\begin{equation}
    \mathrm{sim}_\mathrm{phrase} = -\mathrm{JSD}(\mathbf{a} \parallel \mathbf{a^\prime})
\end{equation}
Here, $\mathbf{a}$ is the attention distribution for $t$, and $\mathbf{a}^\prime$ is the attention distribution for either a positive or a negative sample.
This equation is used to calculate similarities for a given attention head and layer---in SynCLM's implementation, only the last layer is used, and $\mathrm{sim}_\mathrm{phrase}$ is averaged across all attention heads in the last layer before being used with Equation \ref{eq:infonce} for the final loss computation.

The ``tree-guided'' objective proceeds similarly.
A token $t_i$ is sampled which forms the root of the positive tree, $T^+$.
Next, up to three tokens $t^-_1, \ldots, t^-_k$ are sampled such that each $t^-_i$ is not in $T^+$ but is adjacent to a token in $T^+$.
A new negative subtree $T^-_i$ is formed for each $t^-_i$ such that a random non-root token in $T^+$ has been removed from $T^+$ along with its children, and the subtree rooted at $t^-_i$ has taken its place.

We may now define tree similarity as follows, where $T$ is a positive or a negative subtree and $\mathbf{z}_a$ is the hidden representation of token $a$:
\begin{align}
\begin{split}
    \mathrm{sim}_\mathrm{tree} &= \mathrm{cossim}(\mathbf{z}_i, {\textstyle \sum}_{t_j \in T_\mathrm{child}}\:e_{ij}\mathbf{z}_j) \\
    \text{where}\quad T_\mathrm{child} &= T \setminus \{t_i\} \\
    e_{ij} &= \frac{\exp\left(\mathbf{z}_i \cdot \mathbf{z}_j\right)}{\sum_{t_k\in T_\mathrm{child}} \exp\left(\mathbf{z}_i \cdot \mathbf{z}_k\right)}
\end{split}
\end{align}
Informally, we are taking the dot product of the root of the subtree with all other tokens in the subtree, softmaxing this dot product, using it to produce a weighted sum of all hidden representations of tokens in the subtree, and taking the cosine similarity between this weighted sum and the root of the subtree.
The closer these tokens' representations are in the hidden space, the higher this similarity measure will be.
Again, SynCLM uses only the last TLM layer for this objective, and this similarity measure is used with Equation \ref{eq:infonce}.
Note that in a preprocessing step, parses are modified so that subword tokens are syntactic children of the head token of the word they belong to.\footnote{We have elided various implementation details here, such as hyperparameters which control how many sample sets to obtain per input sequence, or maximum token count for a subtree. Please refer to our code or \citet{zhang-etal-2022-syntax}'s code for these details.}

\section{English Experiments}
\label{sec:english}

\begin{table*}[t]
    \centering
    \footnotesize
    
    \begin{tabular}{l|cccc|c}
        Model                        & \np{} & \hqp & \bd & \bd\bm & Avg.\\\hline\hline
        \textsc{$\upmu$b-m}          & 86.79 & 85.60 & 85.81 & 87.83 & 86.51 \\
        \textsc{$\upmu$b-mp}         & 86.89 & 85.36 & 85.91 & 87.73 & 86.47 \\
        \textsc{$\upmu$b-mt}         & 86.51 & 85.83 & 85.93 & 87.10 & 86.34 \\
        \textsc{$\upmu$b-mpt}        & 86.57 & 85.39 & 85.83 & 86.99 & 86.19 \\
        \textsc{$\upmu$b-mpt-sla}    & 86.61 & 85.42 & 85.62 & 86.53 & 86.05 \\\hline
        Avg.                         & 86.67 & 85.52 & 85.82 & 87.23\\
    \end{tabular}
    \caption[English UD parsing results for SynCLM/SLA]{Labeled attachment score (LAS) for English.}
    \label{tab:loreiba_eng_parser_results}
\end{table*}

\paragraph{Parsing}
Parsing results are given in Table \ref{tab:loreiba_eng_parser_results}.
First note that as before, there is little difference in model quality across all the SynCLM conditions, providing more evidence that the SynCLM losses are not helpful for UD parsing.
Next, as could be expected, the model trained with 100M tokens that is half the size of BERT-base performs best.
What is surprising, however, is that of the remaining 3 models, the model with the standard parser performs best.
Since all three of these variants are alike in model hyperparameters, this must be explainable in terms of properties of the three datasets.
It could be that AMALGUM's very deliberate construction from eight genres in equal proportion could have led to serendipitously good performance on the parsing task, but it is impossible to know without further experimentation.

At any rate, whatever the differences in these three variants might be caused by that lies in the data, we still have a firm answer for our most important question: for English UD parsing, SynCLM and SLA methods appear not to be sensitive to data quantity or parse quality.
The latter might be surprising, but it is worth remembering that the authors of these methods designed their algorithms in ways that may mitigate the deleterious effects of lower-quality syntactic parses.
SLA uses windowed syntactic distance (cf.~Equation \ref{eq:windowed} in Appendix \ref{sec:methods}) for the express purpose of accommodating bad parses, and the SynCLM losses place low limits on tree height, which would help in accommodating bad parses since edges at the local, phrase level are often more reliable than edges at the clausal or inter-clausal level.

\begin{table*}[t]
    \centering
    \footnotesize
    \begin{tabular}{l|cccc|c}
        Model                        & \np{} & \hqp & \bd & \bd\bm & Avg.\\\hline\hline
        \textsc{$\upmu$b-m}          & 60.07 & 58.79 & 57.18 & 51.15 & 56.80 \\
        \textsc{$\upmu$b-mp}         & 59.99 & 55.29 & 54.46 & 50.96 & 55.18 \\
        \textsc{$\upmu$b-mt}         & 56.92 & 55.65 & 57.58 & 49.52 & 54.92 \\
        \textsc{$\upmu$b-mpt}        & 61.54 & 59.32 & 55.63 & 49.98 & 56.62 \\
        \textsc{$\upmu$b-mpt-sla}    & 61.49 & 56.05 & 59.51 & 43.90 & 55.24 \\\hline
        Avg.                         & 60.00 & 57.02 & 56.87 & 49.10\\
    \end{tabular}
    \caption[English NER results for SynCLM/SLA]{Span-based F1 score by language and model combination for NER evaluation.}
    \label{tab:loreiba_eng_ner_results}
\end{table*}

\paragraph{NER}
Results on NER are given in Table \ref{tab:loreiba_eng_ner_results}.
Surprisingly, the same half-sized BERT model that was trained on 100M tokens and did best in the parsing evaluation does very poorly in the NER task.
We suspect that this may be due to the fact that larger models can show greater instability in fine-tuning setups \citep{rogers_primer_2020-3}.
As with parsing, we see that the \np{} model performs best among the MicroBERT-sized models, which we ascribe to differences in properties of the pretraining datasets.

What is most interesting in the NER results is that for the two low-data conditions, \np{} and \hqp{}, we see about a 1\% gain in the \textsc{-mpt} condition relative to the MLM-only baseline.
This gain is not seen in the higher-data conditions, where none of the SynCLM combinations lead to a better model except for \mpt{}\bd{}, with a gain of 0.45\% F1.
Complicating this picture, though, is that in the low-data settings, the \textsc{-mp} and \textsc{-mt} variants often underperform relative to the baseline.
Still, these results seem to indicate at least that the SynCLM loss functions may be less effective in improving model quality as quantity of pretraining data increases.
We can see that this holds both for the half-sized BERT model as well as the MicroBERT-sized model, indicating that model size does not matter.

\paragraph{Discussion}
Returning to RQ3, these results indicate that SynCLM and SLA are not especially sensitive to parse quality, and are also not sensitive to model size, but are sensitive to quantity of pretraining data.
As discussed above, the insensitivity to parse quality is understandable, as the dimensions in which a parse may be bad are less relevant for these methods because of the way they use the parse trees.
The sensitivity to pretraining data quantity is intuitive if we consider these two methods as sources of inductive bias: an inductive bias ought to be pushing a model towards learning something that they would have learned if there were more training data available, and so we should expect that if we consider a modification to be an inductive bias, its influence should wane as the quantity of data increases.

In sum, these findings support our conclusion that SynCLM and SLA are at least in some respects well-suited to aid the pretraining of TLMs in low-resource settings, as we have found that even when parse quality is worse than ideal, SynCLM and SLA still perform about as well as when they have the highest quality parses.

\section{Limitations}
The goal of this paper is to make progress towards more effective TLMs for low-resource languages using syntactic inductive bias.
We believe we have presented compelling evidence that two approaches to this problem seem not to be very effective for low-resource languages.
But it is important to point out that we have tested the methods on only 5 languages.
We believe that this forms an informative picture for low-resource languages in general because these languages are quite different from one another along typological and phylogenetic dimensions, but in principle, it is conceivable that other low-resource languages could exhibit behaviors that are very different from the ones we have seen in this paper.
Moreover, we have had to re-implement the methods at the center of this work, and while we have done everything we can to ascertain that these re-implementations have been faithful and without error, tensor programming is error-prone work, and it is not impossible that we may have introduced a bug somewhere which critically affected the experimental results in this work.

\section{Other PrOnto Results}
\label{sec:loreiba_pronto_extra_results}

\begin{table}[h]
    \centering
    \footnotesize
    \begin{tabular}{l|cccc}
                                     & \multicolumn{4}{c}{Proper Noun Subject} \\
        Model                        & An. Grk. & Coptic   & Uyghur & Wolof \\\hline
        \textsc{$\upmu$b-m*}         & 76.32    & 78.76    & 81.30  & 90.36 \\
        \textsc{$\upmu$b-mx*}        & 81.11    & 80.78    & 78.45  & 90.36 \\
        \mbert                       & 81.42    & 75.50    & 80.35  & 91.65 \\\hline
        \textsc{$\upmu$b-m}          & 79.88    & 79.22    & 80.35  & 80.15 \\
        \textsc{$\upmu$b-mp}         & 79.72    & 79.38    & 80.82  & 77.97 \\
        \textsc{$\upmu$b-mt}         & 79.57    & 75.66    & 81.14  & 77.72 \\
        \textsc{$\upmu$b-mpt}        & 76.32    & 79.53    & 77.02  & 77.72 \\\hline
        \textsc{$\upmu$b-mx}         & 81.27    & 81.40    & 80.67  & 81.84 \\
        \textsc{$\upmu$b-mxp}        & 78.79    & 78.91    & 79.71  & 79.42 \\
        \textsc{$\upmu$b-mxt}        & 76.32    & 80.47    & 73.53  & 77.72 \\
        \textsc{$\upmu$b-mxpt}       & 80.80    & 80.16    & 79.40  & 77.72 \\
    \end{tabular}
    \caption{Accuracy by language and model combination for Proper Noun Subject in PrOnto. Scores for \mbert{}, \textsc{$\upmu$b-m*}, and \textsc{$\upmu$b-mx*} are taken from \citet{gessler2023pronto}---the asterisk indicates that the latter two models are not our implementation but the one provided in \citet{gessler-zeldes-2022-microbert}, which is reported in \citet{gessler2023pronto}.}
\end{table}

\begin{table}[h]
    \centering
    \footnotesize
    \begin{tabular}{l|cccc}
                                     & \multicolumn{4}{c}{Sentence Mood} \\
        Model                        & An. Grk. & Coptic   & Uyghur & Wolof \\\hline
        \textsc{$\upmu$b-m*}         & 90.18    & 89.75    & 89.96  & 90.36 \\
        \textsc{$\upmu$b-mx*}        & 91.56    & 89.75    & 90.10  & 90.36 \\
        \mbert                       & 91.70    & 91.55    & 91.23  & 91.65 \\\hline
        \textsc{$\upmu$b-m}          & 91.98    & 91.69    & 91.51  & 90.36 \\
        \textsc{$\upmu$b-mp}         & 90.73    & 91.97    & 91.23  & 89.72 \\
        \textsc{$\upmu$b-mt}         & 90.59    & 90.30    & 89.25  & 90.36 \\
        \textsc{$\upmu$b-mpt}        & 90.73    & 90.30    & 89.96  & 90.36 \\\hline
        \textsc{$\upmu$b-mx}         & 90.59    & 92.24    & 91.80  & 90.58 \\
        \textsc{$\upmu$b-mxp}        & 91.56    & 91.97    & 90.81  & 90.58 \\
        \textsc{$\upmu$b-mxt}        & 91.42    & 90.03    & 89.96  & 90.36 \\
        \textsc{$\upmu$b-mxpt}       & 90.73    & 90.03    & 89.96  & 90.36 \\
    \end{tabular}
    \caption{Accuracy by language and model combination for Sentence Mood in PrOnto. Scores for \mbert{}, \textsc{$\upmu$b-m*}, and \textsc{$\upmu$b-mx*} are taken from \citet{gessler2023pronto}---the asterisk indicates that the latter two models are not our implementation but the one provided in \citet{gessler-zeldes-2022-microbert}, which is reported in \citet{gessler2023pronto}.}
\end{table}

\begin{table}[h]
    \centering
    \footnotesize
    \begin{tabular}{l|cccc}
                                     & \multicolumn{4}{c}{Same Argument Count} \\
        Model                        & An. Grk. & Coptic   & Uyghur & Wolof \\\hline
        \textsc{$\upmu$b-m*}         & 61.80    & 62.70    & 61.78  & 61.05 \\
        \textsc{$\upmu$b-mx*}        & 61.71    & 61.58    & 62.12  & 63.46 \\
        \mbert                       & 50.87    & 51.24    & 50.78  & 54.46 \\\hline
        \textsc{$\upmu$b-m}          & 59.72    & 56.94    & 59.23  & 56.65 \\
        \textsc{$\upmu$b-mp}         & 58.57    & 57.61    & 59.99  & 58.38 \\
        \textsc{$\upmu$b-mt}         & 58.44    & 57.26    & 59.43  & 56.10 \\
        \textsc{$\upmu$b-mpt}        & 53.13    & 56.01    & 59.56  & 56.32 \\\hline
        \textsc{$\upmu$b-mx}         & 57.06    & 55.92    & 60.10  & 56.05 \\
        \textsc{$\upmu$b-mxp}        & 58.60    & 56.18    & 59.87  & 56.21 \\
        \textsc{$\upmu$b-mxt}        & 58.03    & 56.47    & 59.67  & 56.69 \\
        \textsc{$\upmu$b-mxpt}       & 58.36    & 58.01    & 57.88  & 57.54 \\
    \end{tabular}
    \caption{Accuracy by language and model combination for Same Argument Count in PrOnto. Scores for \mbert{}, \textsc{$\upmu$b-m*}, and \textsc{$\upmu$b-mx*} are taken from \citet{gessler2023pronto}---the asterisk indicates that the latter two models are not our implementation but the one provided in \citet{gessler-zeldes-2022-microbert}, which is reported in \citet{gessler2023pronto}.}
\end{table}

\end{document}